\documentclass[letterpaper, 10 pt, conference]{ieeeconf}  

\IEEEoverridecommandlockouts                              

\usepackage{graphics} 
\usepackage{epsfig} 
\usepackage{times} 
\usepackage{amsmath} 
\usepackage{amssymb}  
\usepackage{multirow,booktabs}
\usepackage{color}
\usepackage[table,HTML]{xcolor}
\usepackage{colortbl}
\usepackage{cite}

\title{\LARGE \bf
PanopticSplatting: End-to-End Panoptic Gaussian Splatting
}

\author{Yuxuan Xie$^{1}$, Xuan Yu$^{1}$, Changjian Jiang$^{1}$, Sitong Mao$^{2}$, Shunbo Zhou$^{2}$, Rui Fan$^{3}$, Rong Xiong$^{1}$, Yue Wang$^{1}$
\thanks{$^{1}$Yuxuan Xie, Xuan Yu, Changjian Jiang, Rong Xiong, and Yue Wang are with Zhejiang University, Hangzhou, Zhejiang, China. Yue Wang is the corresponding author {\tt\footnotesize wangyue@iipc.zju.edu.cn}}%
\thanks{$^{2}$Sitong Mao and Shunbo Zhou are with Huawei Cloud Computing Technologies Co., Ltd., Shenzhen, China.}%
\thanks{$^{3}$Rui Fan is with Tongji University, Shanghai, China.}%
}

\begin{document}

\maketitle
\thispagestyle{empty}
\pagestyle{empty}

\begin{abstract}

Open-vocabulary panoptic reconstruction is a challenging task for simultaneous scene reconstruction and understanding.
Recently, methods have been proposed for 3D scene understanding based on Gaussian splatting.
However, these methods are multi-staged, suffering from the accumulated errors and the dependence of hand-designed components.
To streamline the pipeline and achieve global optimization, we propose PanopticSplatting, an end-to-end system for open-vocabulary panoptic reconstruction.
Our method introduces query-guided Gaussian segmentation with local cross attention, lifting 2D instance masks without cross-frame association in an end-to-end way.
The local cross attention within view frustum effectively reduces the training memory, making our model more accessible to large scenes with more Gaussians and objects.
In addition, to address the challenge of noisy labels in 2D pseudo masks, we propose label blending to promote consistent 3D segmentation with less noisy floaters, as well as label warping on 2D predictions which enhances multi-view coherence and segmentation accuracy.
Our method demonstrates strong performances in 3D scene panoptic reconstruction on the ScanNet-V2 and ScanNet++ datasets, compared with both NeRF-based and Gaussian-based panoptic reconstruction methods.
Moreover, PanopticSplatting can be easily generalized to numerous variants of Gaussian splatting, and we demonstrate its robustness on different Gaussian base models.
\end{abstract}

\section{INTRODUCTION}

Open-world panoptic reconstruction is an important task in 3D scene understanding for robotics.
Since the high cost of 3D annotation and the remarkable progress in 2D open-vocabulary segmentation\cite{sam, groundedsam}, most of the existing methods\cite{pvlff, panopticrecon, pr++} lift the ability of 2D VLMs\cite{sam, groundedsam, clip, dino} to 3D for 3D open-vocabulary segmentation. 

3D panoptic reconstruction methods of this type are initially based on NeRF.
As a representative work, Panoptic Lifting\cite{panopticlifting} proposes a multi-view consistent label lifting scheme with linear assignment. 
Further, PanopticRecon++\cite{pr++} uses cross attention to introduce 3D spatial priors, improving the performance of end-to-end panoptic reconstruction.
However, these methods suffer from intensive computation and slow rendering speed due to the implicit representation and random sampling.
Besides, the simultaneous reconstruction and segmentation enables 3D scene editing applications, while NeRF-based methods cannot be easily applied to these tasks, since objects are implicitly encoded in weights of neural networks\cite{decomposingnerf}.

In comparison, 3D Gaussian Splatting has emerged as an explicit representation which shows a significant advantage in rendering speed, and allows editing in a simple way. 
Existing methods\cite{LEGaussian, langsplat, feature3dgs, gaussiangrouping, opengaussian, PLGS} extend it to 3D scene understanding by adding feature attributes to Gaussians.
Some of them\cite{gaussiangrouping, PLGS} propose effective solutions to align and correct 2D machine-generated masks, and lift them to 3D.
Others\cite{LEGaussian, langsplat, feature3dgs, opengaussian} focus on consistent and efficient 2D feature distillation to achieve language-aligned 3D scene understanding.
However, they are all multi-staged, which causes the decrease in performance, including inconsistent label lifting caused by errors in prior stages, and the limited generalization due to the manual components. 



\begin{figure}[t]
    \centering
    \includegraphics[width=\linewidth,keepaspectratio]{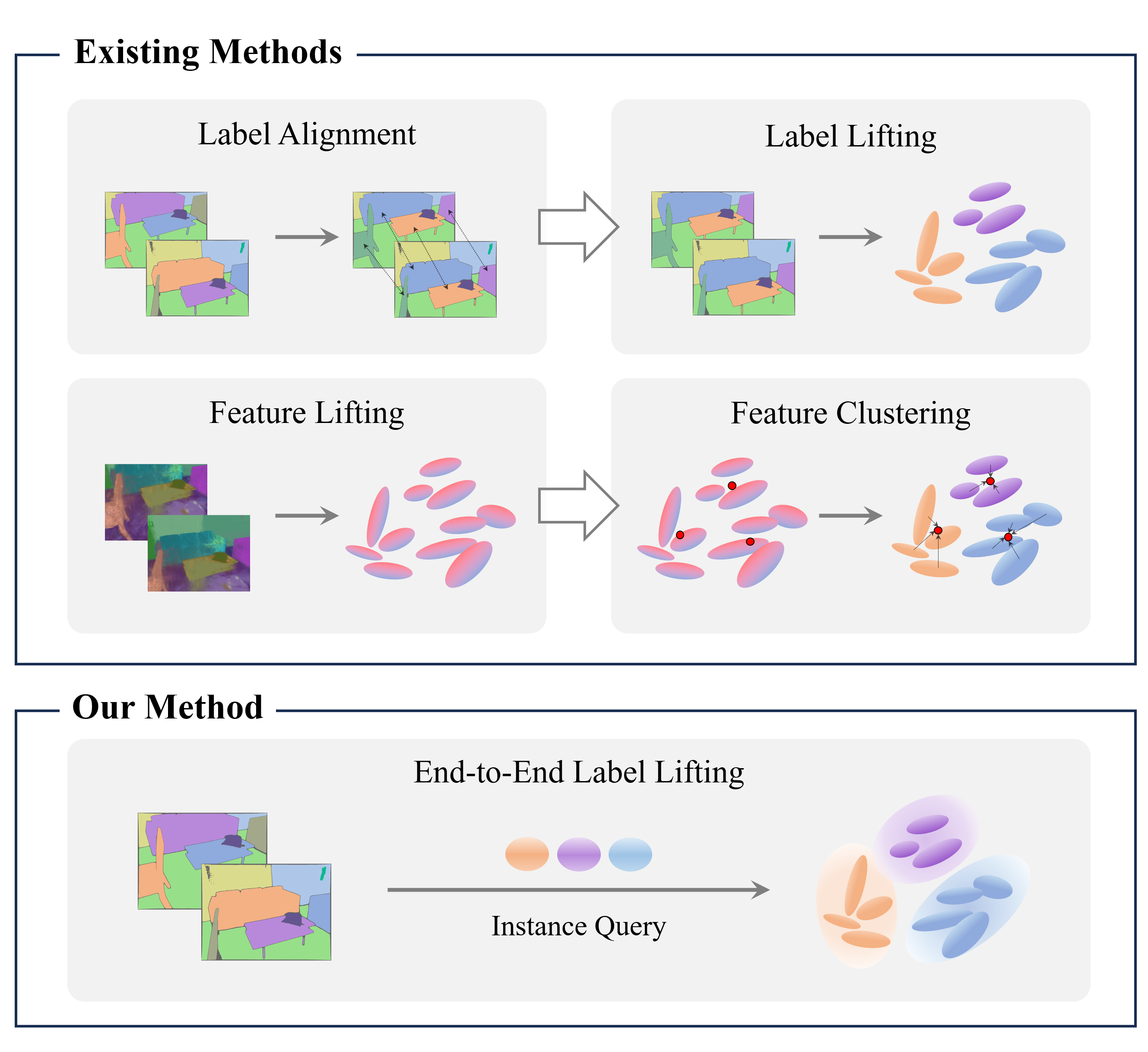}\\
    \vspace{-0.3cm}
    \caption{Existing methods for 3D panoptic reconstruction based on Gaussian splatting have multi-staged procedures. They can be broadly divided into two categories: label-lifting methods decompose the task into 2D mask alignment and label lifting; feature-lifting methods require hand-designed post-processing after feature distillation. Our method introduces instance queries to guide end-to-end panoptic reconstruction.}
    \vspace{-0.3cm}
    \label{fig:first_fig}
\end{figure}





To solve this issue, we aim to learn an end-to-end Gaussian-based model for 3D open-vocabulary panoptic reconstruction.
First, we focus on how the existing methods optimize the instance segmentation to analyze the limitations restricting them from end-to-end optimization.
The Gaussian-based label lifting methods learn fixed instance labels from 2D masks and lack instance optimization in 3D scenes, which prevents them from learning consistent segmentation from misaligned labels.
And the feature lifting methods locate instances based on optimized feature fields, resulting in phased optimization.
Based on the above analysis, we believe that the key to end-to-end learning is simultaneous optimization of feature fields and instance segmentation, which is challenging to implement due to the difficulty in optimizing instances that implicitly exist in the feature field.
To address this challenge, we introduce learnable queries to explicitly model the instances in 3D and propose query-guided Gaussian segmentation for end-to-end panoptic reconstruction.



In this paper, we propose PanopticSplatting, an open-vocabulary panoptic reconstruction method based on Gaussian Splatting, deploying end-to-end training under the supervision of 2D masks from VLMs~\cite{groundedsam}.
Specifically, build upon a set of Gaussians, we add feature attributes to Gaussians to model the semantic and instance fields separately.
To achieve end-to-end instance reconstruction, learnable instance queries are introduced to guide 3D Gaussian segmentation, with cross attention to model the similarity between queries and Gaussians, followed by linear assignment between predicted instances and pseudo masks. 
We save training memory without affecting the performance by constraining the cross attention within view frustum.
To mitigate the effects of 2D noisy labels in semantic reconstruction, we perform label blending instead of common feature blending to enhance 3D segmentation consistency, and propose label warping across views for multi-view consistent segmentation.
In summary, our main contributions are as follows:

\begin{itemize}

\item We propose an end-to-end Gaussian-based method for open-vocabulary panoptic reconstruction by query-guided Gaussian segmentation.
\item We introduce label blending and label warping to mitigate the effects of 2D noisy labels, and reduce the memory cost of query-guided Gaussian segmentation by constraining cross-attention within view frustum.
\item PanopticSplatting shows strong performances in 3D scene panoptic reconstruction compared with existing methods, and we demonstrate its robustness on different Gaussian base models.

\end{itemize}

\begin{figure*}[t]
    \centering
    \includegraphics[width=\linewidth,keepaspectratio]{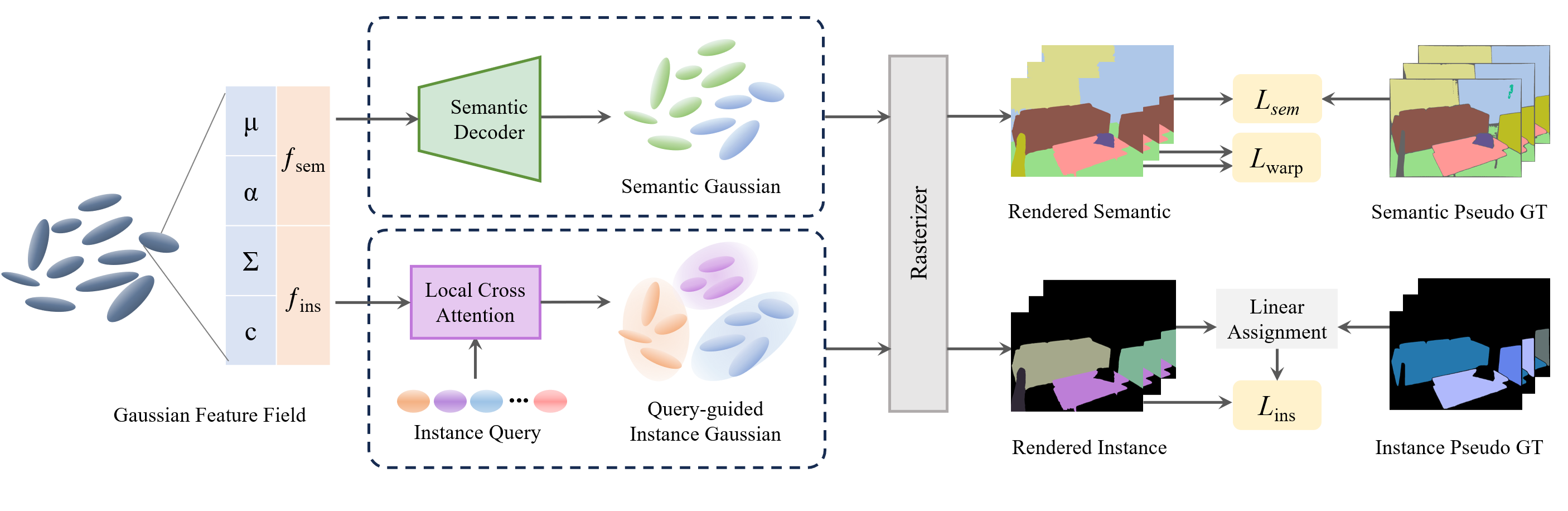}\\
    \vspace{-0.3cm}
    \caption{PanopticSplatting introduces a semantic feature and an instance feature to Gaussians to build semantic and instance feature field. In instance branch, Gaussian-modulated instance queries are introduced to guide Gaussian segmentation through local cross attention. The semantic labels of Gaussians are generated by a simple semantic decoder. Then the labels of Gaussians are rendered to 2D simultaneously. To achieve end-to-end training, linear assignment between 2D pseudo instance masks and predicted labels is performed. We employ the label warping loss on rendered semantic masks.}
    \vspace{-0.3cm}
    \label{fig:Pipeline}
\end{figure*}
\section{RELATED WORKS}

\subsection{Nerf-based Panoptic Reconstruction}

Neural Radiance Field methods encode the scene into a neural network, offering an implicit representation to various properties of 3D scenes, including appearance~\cite{nerf, tensorf, Mip-nerf360, instantngp}, geometry~\cite{neus, neuralrgbd}, semantic~\cite{zhi2021place, tschernezki2022neural, kerr2023lerf}, etc.
Since NeRF effectively connects 2D images and 3D scenes, numerous follow-up\cite{zhi2021place, tschernezki2022neural, panopticlifting} works build neural feature fields to understand 3D scenes through 2D vision knowledge. 
Closer to our work are methods that employ NeRFs to address the problem of 3D panoptic segmentation.
Panoptic Lifting\cite{panopticlifting} proposes a label lifting scheme with a linear assignment between predictions and unaligned instance labels to build a multi-view consistent 3D panoptic representation.
Contrastive Lift\cite{contrastivelift} achieves 3D object segmentation without quantity limitation, through a slow-fast clustering objective function using contrastive learning.
PVLFF\cite{pvlff} also uses contrastive learning to build an instance feature field, and achieves open-vocabulary panoptic segmentation by distilling features of 2D VLMs.
PanopticRecon\cite{panopticrecon} proposes a two-stage method with label propagation and 2D instance association to align 2D masks for 3D panoptic reconstruction. 
PanopticRecon++\cite{pr++} introduces an end-to-end framework for 3D panoptic reconstruction, using Gaussian-modulated instance tokens to guide 3D instance segmentation.

Although these methods perform well in 3D panoptic reconstruction based on NeRF, they suffer from the drawbacks of this implicit and continuous representation. First, they require a significant cost of computation and time for training and rendering, due to the large neural networks and large number of stochastic samples during volumetric ray-marching. Besides, since the implicit and continuous properties of these representations, there are challenges for them to apply to scene editing tasks, which are popular in many areas such as augmented reality and gaming.




\subsection{Gaussian-based Scene Understanding}

3DGS\cite{3dgs} and its numerous variants\cite{2dgs, gaussiansurfels, ligs} optimize a set of differentiable Gaussians to create compact and flexible representation of the 3D scene, providing outstanding results in appearance reconstruction.
In addition, the tile-based rasterizer guarantees the real-time rendering and low memory consumption of Gaussian splatting.
Similar to NeRF-based feature fields, existing methods\cite{LEGaussian, langsplat, feature3dgs, gaussiangrouping, opengaussian} extend it to 3D scene understanding by adding feature attributes for each Gaussian.
LEGaussians\cite{LEGaussian} proposes a language embedded procedure with the quantization scheme and uncertainty modeling to introduce semantics to Gaussians in an efficient and consistent way.
LangSplat\cite{langsplat} designs a scene-wise language autoencoder to reduce feature dimensions and learns hierarchical semantics to address ambiguity in language fields.
Feature 3DGS\cite{feature3dgs} introduces a parallel N-dimensional rasterizer with a speed-up module to distill high-dimensional features.
Gaussian Grouping\cite{gaussiangrouping} uses a tracking method to associate the 2D mask generated by SAM, and then lift the consistent masks to group Gaussians with a 3D Regularization loss.
To enhance 3D point-level scene understanding, OpenGaussian\cite{opengaussian} proposes a multi-stage pipeline that first learns and clusters the instance feature in 3D, followed by 3D-2D feature association.
PLGS\cite{PLGS} introduces semantic anchor points and self-training approach for robust training and generates consistence instance masks through 3D matching.

For 3D panoptic reconstruction task, these existing methods require multi-staged pipelines. The feature lifting\cite{LEGaussian, langsplat, feature3dgs, opengaussian} methods allow open-vocabulary querying and segmentation for a single object through language queries, while when addressing scene panoptic segmentation with objects of uncertain quantity, they need more hand-designed components such as clustering the similar features and setting segmentation thresholds.
The label lifting\cite{gaussiangrouping, PLGS} methods suffer from the 2D masks without association across views, thus they need to align the 2D labels first.
Our method aims to build a Gaussian-based end-to-end panoptic reconstruction system, employing 3D consistency to lift 2D labels in a multi-view unified way.

\section{METHOD}




\subsection{Gaussian-based Feature Representation}

Gaussian Splatting\cite{3dgs} uses 3D Gaussians with geometric and appearance attributes to reconstruct 3D scenes. 
It employs fast differentiable rasterization to rendering RGB images from any view.
During rasterization, 3D Gaussians are first sorted by depth and projected onto the image plane.
Then the color of each pixel can be computed via $\alpha$-blending based on the color attribute:
\begin{equation}
    C = \sum\limits_{i \in \mathcal{N}} c_i \alpha_i' \prod\limits_{j = 1}^{i-1} (1-\alpha_i')
    \label{blending}
\end{equation}
where $\alpha_i'$ represents the influence intensity of Gaussian on this pixel, determined by the opacity of 3D Gaussian and the distribution of projected 2D Gaussian.

To extend 3D Gaussians to panoptic reconstruction, we model the semantic and instance of Gaussians by additionally introducing a semantic feature and an instance feature to them. 
The features are represented by a learnable embedding, denoted as $f_{sem}$ and $f_{ins}$ respectively. 
Note that unlike the anisotropic color attribute, the semantics and instances of Gaussians are consistent across rendering views, thus they can be represented in a view-independent way.
Similar to rendering RGB images, we can obtain 2D semantics and instances by rasterization.

\subsection{Instance Reconstruction}

We lift 2D instance masks to build the 3D instance feature field, while the instance IDs across views are misaligned, which cannot become a consistent supervision for scene-level instance segmentation. Thus, to associate 2D labels by using 3D consistency, we introduce instance queries to guide the instance segmentation on the 3D feature field, ensuring consistent instance labels in 3D.


{\bf Query-Guided Gaussian Segmentation.}
One of the common paradigms for end-to-end instance segmentation is to use object queries to guide object detection\cite{detr} and segmentation\cite{maskformer}. 
Through the interaction between scene features and query features, queries learn object features and guide the scene instance feature clustering.
Then, the constituent units of the scene, such as 2D pixels or 3D points, can be segmented through the similarity with queries.

In 3D scene segmentation, previous work\cite{pr++} uses spatial distance weighted attention map to model this similarity between 3D points and queries which are encoded as 3D Gaussian distribution. We also adopt this similarity representation, using distance-aware cross attention between instance queries and instance feature of Gaussians to segment Gaussians in 3D.

Considering that the scene Gaussians are far smaller in size than the Gaussian-modulated queries, we simplify the scene Gaussians into points when performing cross attention with queries. 
First, the similarity between query feature $f_q$ and Gaussian instance feature $f_{ins}$ is defined as:
\begin{equation}
    {S}(f_q,f_{ins}) = {sigmoid}(f_q^T f_{ins})
\end{equation}

Besides, since the instance queries are Gaussian distributed, the distance from a 3D point to the query can be described by the probability density. The distance between query $q$ and scene Gaussian $g$ is defined as:
\begin{equation}
    D(p_q,p_g) = \frac{\varphi(p_g)}{\varphi(p_q)}
    \label{eq:wight}
\end{equation}
where $\varphi(\cdot)$ is the probability density function of query's Gaussian distribution. $p_q$ and $p_g$ are the center point coordinates of the instance query and scene Gaussian, respectively.
Thus, the attention map is as follows:
\begin{equation}
    {A}(q,g) = {S}(f_q,f_{ins}) {D}(p_q,p_g)
\label{gsours}
\end{equation}

The instance label of Gaussian $g$ is defined as:
\begin{equation}
    l_{ins}(g)= softmax(\left[{A}(q_1,g) \ldots {A}(q_i,g) \ldots {A}(q_N,g) \right])
\end{equation}
where $N$ is the number of queries. 

Then, similar to (\ref{blending}), the instance label of pixels can be computed via $\alpha$-blending the Gaussian labels:
\begin{equation}
    I = \sum\limits_{i \in \mathcal{N}} l_{ins}(i) \alpha_i' \prod\limits_{j = 1}^{i-1} (1-\alpha_i')
    \label{eq:ins_blending}
\end{equation}




{\bf Local Cross Attention.}
Considering the large number of Gaussians in the entire scene, cross attention between Gaussians and queries requires high computational cost, thus we introduce local cross attention to limit the usage of training memory and shorten the training time.
For each training view, only Gaussians within the view frustum will update via back propagation during the training process. Thus, we perform local cross attention, which limits the cross attention within view frustum.

Following the previous work\cite{3dgs}, Gaussians within the view frustum are defined as Gaussians with a 99$\%$ confidence interval intersecting the view frustum. 
We implement local cross attention in CUDA kernel. To be specific, after filtering out the Gaussians within the view frustum, we launch one thread for each selected Gaussian to compute cross attention in parallel. 
Thus, the attention map is obtained efficiently as the instance feature of 3D valid Gaussians. Then, as shown in (\ref{eq:ins_blending}), we perform tile-based rasterization to obtain 2D instance maps.
This strategy effectively reduces the training memory compared with global cross attention, especially when the number of scene Gaussians increases, making our model more accessible to large scenes with more Gaussians.

\begin{figure}[t]
    \centering
    \includegraphics[width=\linewidth,keepaspectratio]{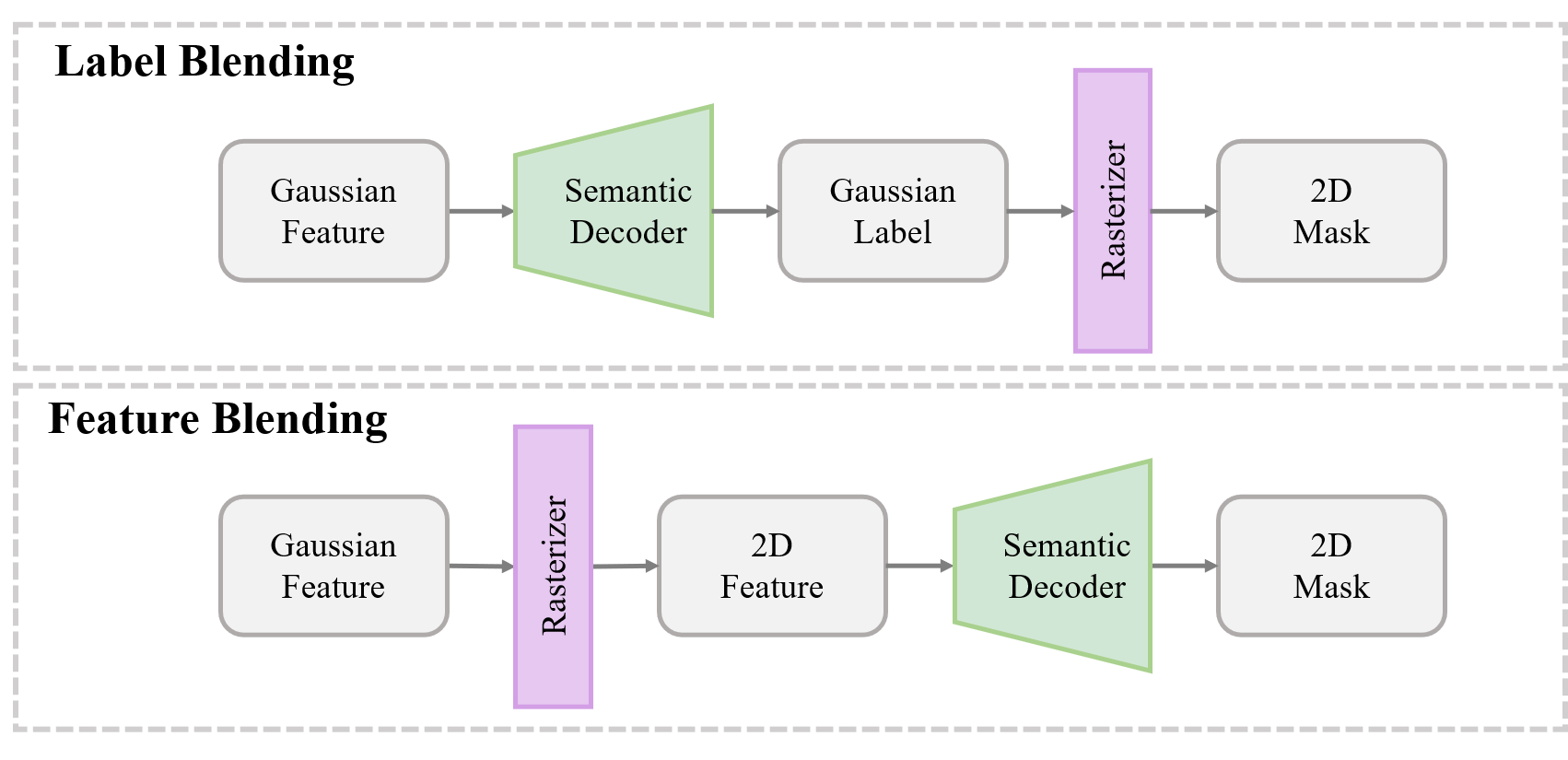}\\
    \vspace{-0.3cm}
    \caption{The pipelines of label blending and feature blending.}
    \vspace{-0.3cm}
    \label{fig:label_blending}
\end{figure}

\subsection{Semantic Reconstruction}

The Gaussian semantic field can be learned through lifting the 2D semantic labels, while we find that the predictions struggle with multi-view semantic inconsistency due to the noisy labels in 2D pseudo masks. To address this issue, we propose label blending to promote consistent 3D segmentation compared with feature blending, and introduce label warping loss on 2D predictions to enhance segmentation accuracy and correct errors in pseudo masks.

{\bf Label Blending.} 
During the rasterization, different from feature blending in previous work\cite{gaussiangrouping}, we obtain the 2D semantic mask by $\alpha$-blending the semantic label of Gaussians instead of semantic feature. Specifically, the Gaussian semantic feature $f_{sem}$ is first decoded by a MLP and a softmax layer to gain the semantic label of the Gaussian, donated as $l_{sem} \in \mathbb{R}^{N_c}$, where $N_c$ is the number of categories. Then, the 2D semantic mask can be computed via label blending:
\begin{equation}
    S = \sum\limits_{i \in \mathcal{N}} l_{sem}(i) \alpha_i' \prod\limits_{j = 1}^{i-1} (1-\alpha_i')
\end{equation}

Fig. \ref{fig:label_blending} shows the pipelines of label blending and feature blending. In label blending, Gaussians are first classified in 3D according to Gaussian semantic features, followed by splatting to 2D and blending. While feature blending pipeline learns a semantic head in 2D, whose strong fitting ability weakens the categories of Gaussians. Thus, label blending emphasizes the 3D Gaussian segmentation which enhances the 3D consistency to ease the impact of 2D noisy labels.
Besides, in label blending pipeline, the softmax layer normalizes Gaussian feature before blending, avoiding that the 2D predictions are dominated by Gaussians with considerable values while other Gaussians with wrong labels along this direction may not be punished.
In the ablation study, a detailed comparison is provided between semantic label blending and feature blending.

{\bf Label Warping Loss.}
Due to the lack of correlation between discrete 3D Gaussians, the feature field based on Gaussian has weaker smoothness compared with NeRF-based methods, which is manifested as the semantic inconsistency across views in 2D predictions under the impact of noisy labels.
Thus, to enhance the association and consistency between multi-view label predictions and correct the few semantic errors in 2D, we propose a per-pixel label warping loss on 2D semantic predictions.

For a pixel in frame $m$, denoted as $r_m$, we first unproject it to 3D using depth and intrinsic to get the pointcloud $p_m$ in the coordinate system of frame $m$:
\begin{equation}
    \begin{bmatrix} p_m \\ 1 \end{bmatrix}
    = D(r_m)K^{-1}
    \begin{bmatrix} r_m \\ 1 \end{bmatrix}
    \label{unproject}
\end{equation}
where $D(r_m)$ donates the depth on pixel $r_m$ and K donates the intrinsic of camera.
Then we transfer the pointcloud to the coordinate system of the adjacent frame $n$:
\begin{equation}
    \begin{bmatrix} p_{m \to n} \\ 1 \end{bmatrix}
    = T_nT_m^{-1}
    \begin{bmatrix} p_m \\ 1 \end{bmatrix}
\end{equation}
where $T_m$ and $T_n$ denote the extrinsic matrix of frame $m$ and $n$.
Then the projected pixel $r_{m \to n}$ in the nearby frame $n$ can be generated by projecting $p_{m \to n}$ to frame $n$ as the reverse process of (\ref{unproject}).
The label warping loss is then defined as:
\begin{equation}
    \mathcal{L}_{warp}(r_m) = \sum\limits_{n \in \mathcal{K}_m} \|M_{sem}(r_m)-M_{sem}(r_{m \to n})\|
    \label{eq:blending}
\end{equation}
where $\mathcal{K}_m$ denotes the adjacent frame list for the current frame $m$. We mask out the pixels that are projected outside the image boundary of frame $n$.

\begin{table*}[]
\caption{Panoptic Segmentation quality using different methods}
\centering
\setlength{\tabcolsep}{5pt}
\renewcommand\arraystretch{1.2}
\resizebox{\linewidth}{!}{
\begin{tabular}{lccc|cc|cc|ccc|cc|cc}
\toprule
\multirow{2}{*}{\textbf{Method}} & \multicolumn{7}{c}{ScanNet-V2} & \multicolumn{7}{c}{ScanNet++} \\
\cmidrule(r){2-8} \cmidrule(r){9-15}
 & PQ & SQ & RQ & mIoU & mAcc & mCov & mW-Cov & PQ & SQ & RQ & mIoU & mAcc & mCov & mW-Cov \\
\midrule
Panoptic Lifting 
    & 57.86 & 61.96 & 85.31 & 67.91 & 78.59 & 45.88 & 59.93 
    & 71.14 & 77.48 & 88.14 & 81.34 & \textbf{89.67} & 56.17 & 68.51 \\
Contrastive Lift 
    & 37.35 & 41.91 & 57.60 & 64.77 & 75.80 & 13.21 & 23.26
    & 47.58 & 57.23 & 65.81 & 81.09 & 89.30 & 27.39 & 36.51 \\
PVLFF 
    & 30.11 & 51.71 & 44.43 & 55.41 & 63.96 & 45.75 & 48.41
    & 52.24 & 66.86 & 65.56 & 62.53 & 70.31 & 67.95 & 75.47 \\
PanopticRecon
    & 63.70 & 64.81 & 81.17 & 68.62 & 80.87 & 66.58 & 77.84
    & 68.29 & 77.01 & 85.05 & 77.75 & 87.08 & 51.34 & 62.79 \\
\midrule
Gaussian Grouping
    & 43.75 & 50.63 & 72.68 & 58.05 & 68.68 & 52.70 & 58.10
    & 33.10 & 40.60 & 67.27 & 59.53 & 68.13 & 29.83 & 36.83\\
OpenGaussian 
    & 48.73 & 51.48 & 88.10 & 54.05 & 68.43 & 44.43 & 49.60 
    & 51.03 & 56.93 & 85.73 & 61.80 & 73.97 & 50.00 & 51.02  \\
\textbf{Ours} 
    & \textbf{74.75} 
    & \textbf{74.75}
    & \textbf{100.0}
    & \textbf{74.95}
    & \textbf{83.70}
    & \textbf{73.18}
    & \textbf{79.63}
    & \textbf{77.73}
    & \textbf{82.70}
    & \textbf{93.60}
    & \textbf{81.90}
    & {89.50}
    & \textbf{74.73}
    & \textbf{78.03}\\
\bottomrule
\end{tabular}
}
\label{tab:Comparative_baseline}
\end{table*}

\subsection{End-to-End Training}


{\bf Instance Assignment.}
To achieve end-to-end 2D instance supervision, we use the Hungarian Algorithm for linear assignment between pseudo instance groundtruth and predicted labels on each frame.

Specifically, we first calculate the matching cost between the binary masks in a pair of pseudo GT and predicted masks. Since the number of instance queries is larger than GT, each GT instance will be distributed a predicted mask in Hungarian Algorithm. We get the the optimal assignment that builds the connection between GT instances and predicted ones by minimizing the total assignment cost.


{\bf Loss Function.}
After instance assignment, we use dice loss and binary cross-entropy loss between a pair of instance prediction $M_{ins}$ and aligned instance GT $M_{ins}^{gt}$ to supervise the instance branch. The instance loss $\mathcal{L}_{ins}$ is defined as:
\begin{equation}
\mathcal{L}_{ins} = \mathcal{L}_{dice}(M_{ins}, M_{ins}^{gt}) + \mathcal{L}_{bce}(M_{ins}, M_{ins}^{gt})
\label{eq:instance_loss}
\end{equation}

The semantic branch is supervised by a cross-entropy loss between rendered semantic $M_{sem}$ and semantic GT $M_{sem}^{gt}$:
\begin{equation}
    \mathcal{L}_{sem} = \mathcal{L}_{ce}(M_{sem}, M_{sem}^{gt})
\end{equation}



The image rendering is supervised with the groundtruth images by a combination of L1 loss and SSIM loss. With the rendered image denoted as $I$ and the groundtruth donated as $I^{gt}$, the rendering loss is defined as:
\begin{equation}
    \mathcal{L}_{rgb} = (1-\lambda_{SSIM})\mathcal{L}_1(I, I^{gt}) + \lambda_{SSIM}\mathcal{L}_{SSIM}(I, I^{gt})
\end{equation}

Combined with the label warping loss in (\ref{eq:blending}), our total loss $\mathcal{L}$ is:
\begin{equation}
    \mathcal{L} = \mathcal{L}_{rgb}+\mathcal{L}_{ins}+\mathcal{L}_{sem}+\mathcal{L}_{warp}
\end{equation}

\begin{figure*}[t]
    \centering
    \includegraphics[width=\linewidth,keepaspectratio]{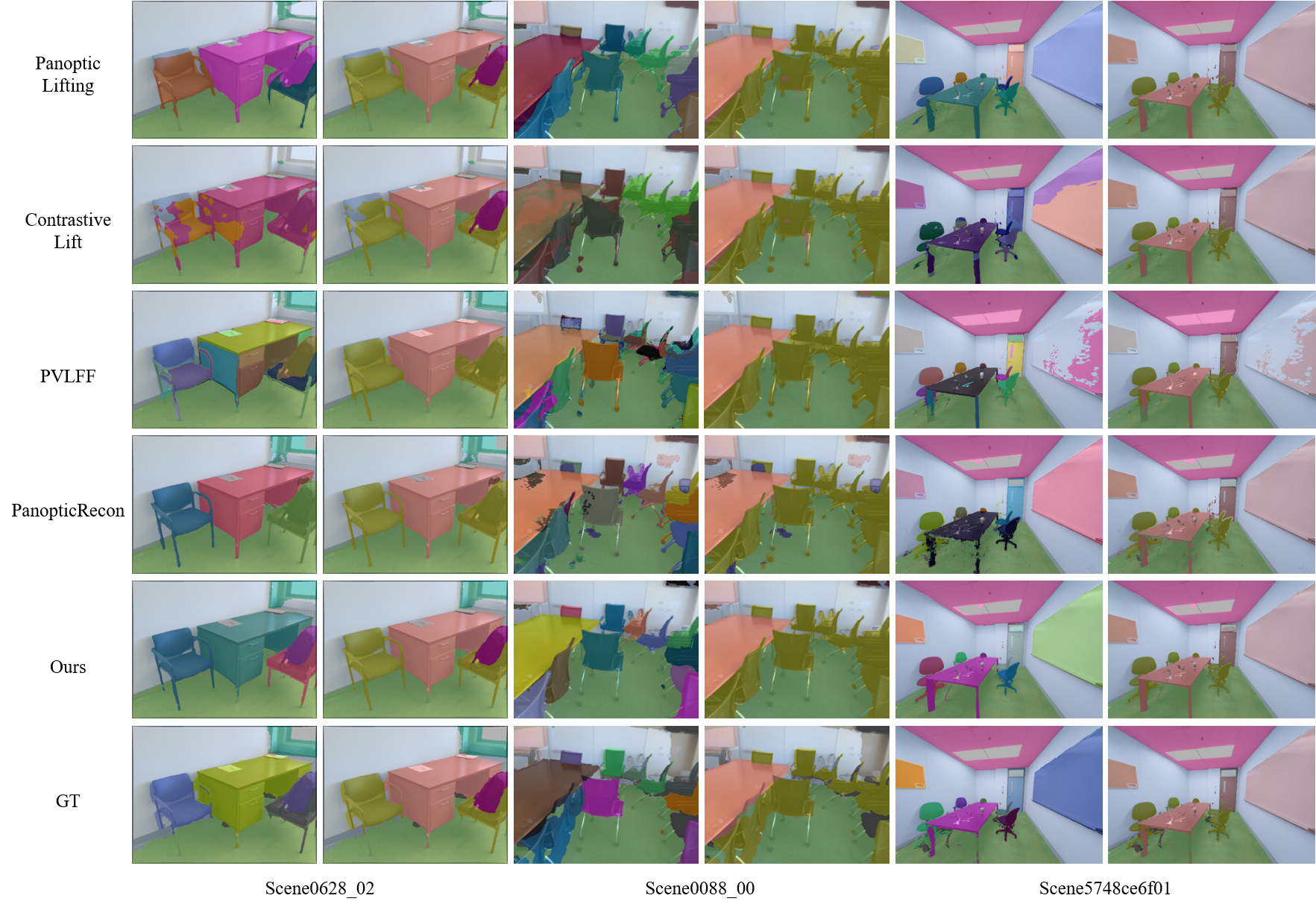}\\
    \vspace{-0.3cm}
    \caption{Comparison of the quality of panoptic segmentation and semantic segmentation of NeRF-based methods on ScanNet-V2 and ScanNet++.}
    \vspace{-0.3cm}
    \label{fig:NeRF-based}
\end{figure*}

\section{EXPERIMENTS}

We validate the performance of our method in real-world datasets, comparing with related works based on NeRF and Gaussian.
We also deploy our method on several representative Gaussian models to illustrate the robustness to Gaussian base models.
Besides, We conduct ablation studies to demonstrate the effectiveness of network components.

\subsection{Setup}

{\bf Datasets.} 
We leverage two real-world datasets for evaluating our method: ScanNet-V2\cite{scannet} and ScanNet++\cite{scannet++}. ScanNet-V2 is an RGB-D dataset that contains diverse real-world indoor scenes with images, labels and reconstructed geometry, which makes it suitable for 3D reconstruction and segmentation. ScanNet++ provides high-resolution 3D scenes with fine annotations, crucial for novel view synthesis and 3D scene understanding tasks. We use 4 scenes in ScanNet-V2 and 3 scenes in ScanNet++ for evaluation.

{\bf Baselines.}
We select several baselines for 3D panoptic segmentation based on Nerf and Gaussian. 
Panoptic Lifting\cite{panopticlifting}, Contrastive Lift\cite{contrastivelift}, PVLFF\cite{pvlff}, and PanopticRecon\cite{panopticrecon} are representative scene segmentation methods based on NeRF. Gaussian Grouping\cite{gaussiangrouping} and OpenGaussian\cite{opengaussian} are Gaussian-based methods that lift 2D vision knowledge for scene understanding. We fairly use the same 2D labels with our method to supervise comparison methods except PVLFF and Gaussian Grouping, for which we use their default 2D mask generating methods. For OpenGaussian, we add post-processing on its binary mask predictions to generate panoptic masks.

{\bf Evaluation Metrics.} 
We evaluate the scene-level segmentation metrics in 2D predictions. Compared with image-level segmentation, scene-level segmentation requires multi-view consistent identities for the same instance, which ensuring the segmentation consistency in 3D.
Following previous work\cite{panopticlifting}, we evaluate panoptic quality (PQ), semantic quality (SQ) and recognition quality (RQ) for 2D panoptic segmentation, together with mIoU, mAcc for semantic segmentation and mCov, mWCov for instance segmentation. 


{\bf Implementation.}
We implement Grounded-SAM\cite{groundedsam} to generate 2D semantic and instance masks as supervision, with the IDs of the same instance in multi-view masks are inconsistent.
In comparative studies with baselines and ablation studies, we deploy our method on LI-GS\cite{ligs}, a scene reconstruction model based on Gaussian surfels.
The semantic feature of Gaussians has 16 dimensions, and the instance feature of Gaussians and query feature have 32 dimensions.
All experiments run on a single A6000 GPU.

\subsection{Comparative Study} 
\begin{figure*}[t]
    \centering
    \includegraphics[width=\linewidth,keepaspectratio]{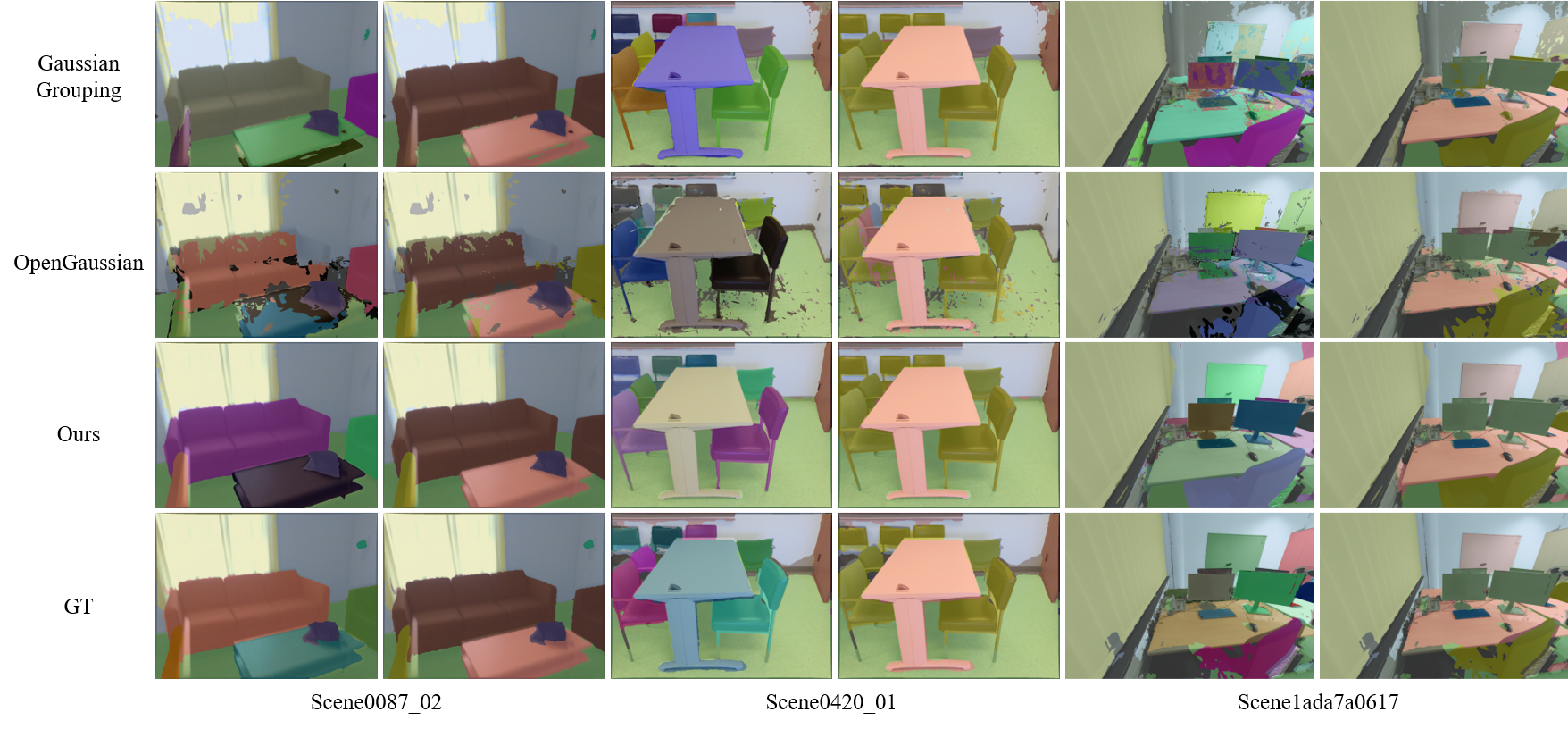}\\
    \vspace{-0.3cm}
    \caption{Comparison of the quality of panoptic segmentation and semantic segmentation of Gaussian-based methods on ScanNet-V2 and ScanNet++.}
    \vspace{-0.3cm}
    \label{fig:Gaussian-based}
\end{figure*}

{\bf Comparative Study with Baselines.}
The quantitative results on ScanNet-V2 and ScanNet++ datasets are shown in Tab.\ref{tab:Comparative_baseline}.
Fig.~\ref{fig:NeRF-based} and Fig.~\ref{fig:Gaussian-based} present the visualization of the comparative results with NeRF-based and Gaussian-based methods on the two datasets.

Our method significantly outperforms both the NeRF-based and Gaussian-based methods. We attribute the segmentation errors of the comparative methods to the following reasons, and prove the superiority of PanopticSplatting.

First, the methods with multi stages, such as PanopticRecon and Gaussian Grouping, suffer from accumulated errors.
As shown in "Scene0628\_02" (Fig.~\ref{fig:NeRF-based}), PanopticRecon fails to segment the "bag" on the chair, which is caused by the under segmentation in its first phase. In Gaussian Grouping, 2D consistent labels generated by tracking is not optimistic in indoor scenes, especially in the ScanNet++ dataset where there is a significant view change between adjacent frames. Thus, the errors in label association lead to the failure of segmentation, as shown in "Scene1ada7a0617" (Fig.~\ref{fig:Gaussian-based}).
Compared with them, our method propose an end-to-end pipeline to effectively avoid accumulated errors and achieve global optimization.

Besides, the non-unique label of 3D instance, manifested as different predictions of the same instance across views, leads to the low scene-level metrics of PanopticLifting and OpenGaussian. Compared with their implicit scene instance representation, we leverage query tokens to explicitly model the 3D instances, learning both the instance feature and 3D spatial priors, which promotes that the masks of one instance on different views correspond to a unique 3D instance.

The feature-lifting methods struggle with accurately separate objects with similar features. For example, Contrastive Lift fails to segment the chairs in the "Scene0088\_00" (Fig.~\ref{fig:NeRF-based}), since the chairs share similar features.
OpenGaussian uses two-level discretization to add 3D spatial priors, while the spatial priors introduced by simple clustering are coarse.
In contrast, our method introduces deformable and learnable instance queries to learn more precision 3D spatial priors, and deploys distance-weighted similarity, which effectively guide spatial-aware Gaussian segmentation.

In addition, to address noisy labels, we enhance the multi-view semantic segmentation consistency by label blending and label warping, which effectively improve segmentation accuracy of stuff categories. As shown in "Scene0420\_01" (Fig.~\ref{fig:Gaussian-based}), only our method accurately segment the "door" among the Gaussian-based methods.

{\bf Study on Model Robustness.} 
We verify the robustness of our model on typical and novel Gaussian bases, including 3DGS\cite{3dgs}, 2DGS\cite{2dgs}, and LI-GS\cite{ligs}.
Tab.\ref{tab:Comparative_basemodel} shows the results on three bases. As we can see, our method achieves good performance on these base models, which demonstrates that it can adapt to different Gaussian models and handle scene panoptic reconstruction methods on different Gaussian representations.
In addition, LI-GS based model achieves the best segmentation performance with the fewest number of Gaussians, mainly benefiting from its excellent geometric reconstruction ability with less floaters. Therefore, we use LI-GS as the base model in other experiments.

\begin{table}[]
\caption{Study on Model Robustness on ScanNet-V2 dataset}
\vspace{-0.2cm}
\centering
\setlength{\tabcolsep}{4pt}
\renewcommand\arraystretch{1.2}
\resizebox{\linewidth}{!}{
\begin{tabular}{lccc|cc|cc}
\toprule
  & PQ & SQ & RQ & mIoU & mAcc & mCov & mW-Cov \\
\midrule
3DGS
    & 73.83 & 74.08 & 99.33 & \textbf{75.08} & 83.68 & 72.20 & 76.43  \\
2DGS
    & 73.30 & 73.45 & 99.93 & 73.75 & 82.58 & 71.28 & 75.98  \\
LI-GS  
    & \textbf{74.75} & \textbf{74.75} & \textbf{100.0} & 74.95 & \textbf{83.70} & \textbf{73.18} & \textbf{79.63} \\
\bottomrule
\end{tabular}
}
\label{tab:Comparative_basemodel}
\end{table}

\begin{table}[]
\caption{Ablation Studies on Label Blending and Label Warping}
\vspace{-0.2cm}
\centering
\setlength{\tabcolsep}{4pt}
\renewcommand\arraystretch{1.2}
\resizebox{\linewidth}{!}{
\begin{tabular}{lccc|cc|cc}
\toprule
  & PQ & SQ & RQ & mIoU & mAcc & mCov & mW-Cov \\
\midrule
FB 
    & 72.43 & 73.73 & 98.60 & 73.13 & 82.95 & 71.03 & 77.90 \\
LB 
    & 73.50 & 74.60 & 98.68 & 74.38 & 82.73 & 73.00 & 79.40 \\
LB+WP  
    & \textbf{74.75} & \textbf{74.75} & \textbf{100.0} 
    & \textbf{74.95} & \textbf{83.70} 
    & \textbf{73.18} & \textbf{79.63} \\
\bottomrule
\end{tabular}
}
\label{tab:Ablation_1}
\vspace{-0.5cm}
\end{table}

\subsection{Ablation Study}

We conduct ablation experiments of the designs in our model to validate the effectiveness of our method. All ablation experiments are performed on the ScanNet-V2 dataset.

{\bf Label Blending.} 
We compare the performance of label blending and feature blending in semantic branch to validate the advantage of label blending. 
As shown in Tab.\ref{tab:Ablation_1}, the label blending (LB) outperforms the feature blending (FB) on both semantic and instance segmentation.

Specifically, the panoptic metrics show that when the recognition quality (RQ) is comparable, label blending achieves better semantic quality (SQ) than feature blending, which means label blending predicts more precise masks for objects.
This is mainly attributed to the 3D segmentation consistency enhanced by semantic segmentation on 3D Gaussians instead of 2D features, and less noisy floaters with normalization on 3D Gaussians.

{\bf Label Warping Loss.}
We study the influence of the label warping loss on the segmentation accuracy. As shown in Tab.\ref{tab:Ablation_1}, adding label warping loss (WP) achieves effective improvement of segmentation metrics, especially on semantic segmentation accuracy, which demonstrates that the model profits from multi-view consistency to alleviate the impact of noisy labels.

\begin{figure}[t]
    \centering
    \includegraphics[width=\linewidth,keepaspectratio]{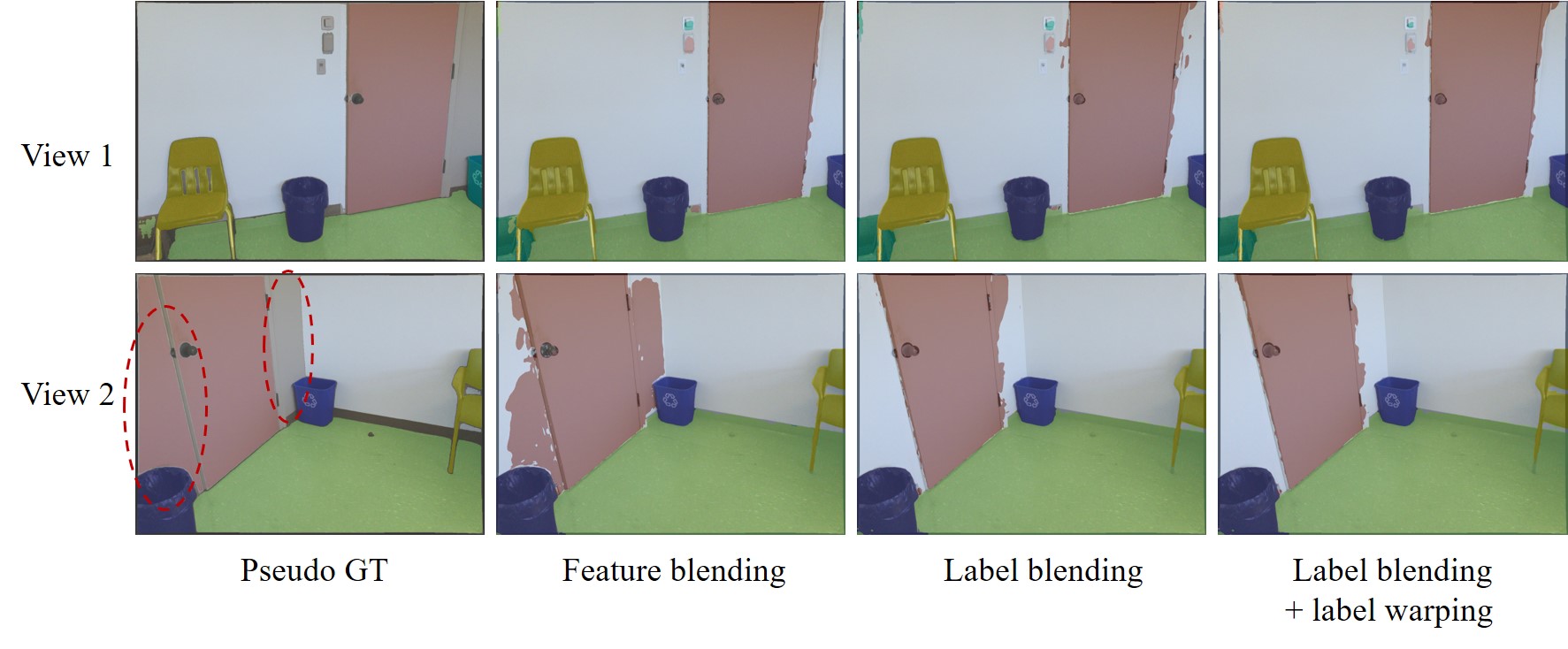}\\
    \vspace{-0.3cm}
    \caption{Ablation on label blending and label warping loss. The noisy labels in View 2 of pseudo GT lead to obvious multi-view inconsistency in feature blending. The label blending effectively addresses this by enhancing 3D segmentation consistency, and the warping loss further correct the incorrect labels through multi-view consistency.}
    \vspace{-0.3cm}
    \label{fig:sem_ablation}
\end{figure}

Fig. \ref{fig:sem_ablation} shows the ablation on label blending and label warping loss. As we can see, noisy labels in View 2 of pseudo GT lead to obvious multi-view inconsistency in the predictions of feature blending. The label blending effectively addresses the inconsistency by enhancing 3D consistency of semantic segmentation, and the warping loss further corrects the incorrect labels.

\begin{table}[]
\caption{Ablation Study on Local Cross Attention}
\vspace{-0.2cm}
\centering
\renewcommand\arraystretch{1.2}
\resizebox{\linewidth}{!}{
\begin{tabular}{lcc|ccc|c}
\toprule
 & Base model & $N_{gs}$ & PQ & mIoU & mCov & Memory \\
\midrule
\multirow{2}{*}{Global CA} &
LI-GS & 296K
    & 74.60 & 74.87 & \textbf{73.26} & 18.59G\\
 & 2DGS & 569K
    & 73.56 & 73.80 & 71.13 & 28.21G\\
\multirow{2}{*}{Local CA} &
LI-GS & 296K
    & \textbf{74.75} & \textbf{74.95} & 73.18 & \textbf{12.21G}\\
 & 2DGS & 569K
    & 73.30 & 73.75 & 71.28 & \textbf{13.39G}\\
\bottomrule
\end{tabular}
}
\label{tab:Ablation_2}
\vspace{-0.5cm}
\end{table}



{\bf Local Cross Attention.} 
We compare the performance and training memory cost between using local cross attention and global cross attention based on LI-GS and 2DGS.
As shown in Tab.\ref{tab:Ablation_2}, the local cross attention between Gaussians and queries effectively decreases the memory cost without affecting the performance.
In particular, when the number of scene Gaussians $N_{gs}$ increases, the memory of local cross attention increases significantly less than that of global, which makes our model fit to large scenes with more Gaussians.


\section{CONCLUSION}

In this paper, we propose PanopticSplatting, an end-to-end panoptic reconstruction method based on Gaussian splatting. Our method first builds Gaussian semantic and instance fields by adding features to each Gaussian. Then query-guided 3D Gaussian segmentation and linear assignment between instance predictions and pseudo GT ensure the end-to-end instance reconstruction. To address the challenge of noisy labels in 2D pseudo masks, we further introduce label blending and label warping to promote consistent segmentation and enhance segmentation accuracy. PanopticSplatting demonstrates strong performance in numerous scenes and good generalization on different Gaussian base models.
\bibliographystyle{IEEEtran}
\bibliography{IEEEabrv,bibliography}


\end{document}